\documentclass[12pt]{article}

\usepackage[a4paper,total={18cm,27cm}]{geometry}
\usepackage{graphicx}
\usepackage{authblk}
\usepackage{amsmath}
\usepackage{amssymb}
\usepackage[utf8]{inputenc}
\usepackage{hyperref}
\usepackage{verbatim}
\usepackage{color,soul}
\usepackage{caption}
\usepackage{xcolor}
\usepackage{framed}
\usepackage{array}
\usepackage{cite}
\colorlet{shadecolor}{yellow!20}

\newcommand{\NI}{\vspace{0.2cm}\noindent}

\begin{document}



\title{Nonlinear Neural Dynamics and Classification Accuracy\\ in Reservoir Computing}


\author[1]{Claus Metzner}
\author[1,2]{Achim Schilling}
\author[1]{Andreas Maier}
\author[1,2]{Patrick Krauss}

\affil[1]{\small Cognitive Computational Neuroscience Group, Pattern Recognition Lab, Friedrich-Alexander-University Erlangen-Nürnberg (FAU), Germany}
\affil[2]{\small Neuroscience Lab, University Hospital Erlangen, Germany}

\maketitle


\vspace{0.4cm}
\begin{abstract}

Reservoir computing - information processing based on untrained recurrent neural networks with random connections - is expected to depend on the nonlinear properties of the neurons and the resulting oscillatory, chaotic, or fixpoint dynamics of the network. However, the required degree of nonlinearity and the range of suitable dynamical regimes for a given task are not fully understood. To clarify these questions, we study the accuracy of a reservoir computer in artificial classification tasks of varying complexity, while tuning the neuron's degree of nonlinearity and the reservoir's dynamical regime. We find that, even for activation functions with extremely reduced nonlinearity, weak recurrent interactions and small input signals, the reservoir is able to compute useful representations, detectable only in higher order principal components, that render complex classificiation tasks linearly separable for the readout layer. When increasing the recurrent coupling, the reservoir develops spontaneous dynamical behavior. Nevertheless, the input-related computations can 'ride on top'  of oscillatory or fixpoint attractors without much loss of accuracy, whereas chaotic dynamics reduces task performance more drastically. By tuning the system through the full range of dynamical phases, we find that the accuracy peaks both at the oscillatory/chaotic and at the chaotic/fixpoint phase boundaries, thus supporting the 'edge of chaos' hypothesis. Our results, in particular the robust weakly nonlinear operating regime, may offer new perspectives both for technical and biological neural networks with random connectivity.
\end{abstract}

\section{Introduction}

Deep learning models, which over the past decades have seen enormous progress  \cite{lecun2015deep,alzubaidi2021review}, including present large language models \cite{min2023recent}, are dominated by artificial neural networks with a feedforward architecture, so that information passes sequentially from input to output layers. In contrast, Recurrent Neural Networks (RNNs) rely on feedback connections and thus function as autonomous dynamical systems \cite{maheswaranathan2019universality}, capable of maintaining neural activity even without ongoing external inputs. 

\NI Certain 'universal' properties of RNNs, such as their ability to approximate arbitrary functions \cite{schafer2006recurrent} or dynamical systems \cite{aguiar2023universal}, and other unique strengths of these systems have driven a substantial research interest into the detailed features of RNNs, for example their ability to retain the information of sequential input time series over extended periods \cite{jaeger2001echo,schuecker2018optimal,busing2010connectivity,dambre2012information,wallace2013randomly,gonon2021fading}, or their learning of effective representations by dynamically balancing the compression and expansion of information \cite{farrell2022gradient}. The control of RNN dynamics is another key research area, including studies that investigate how external and internal noise can influence network behavior \cite{rajan2010stimulus,jaeger2014controlling,haviv2019understanding,molgedey1992suppressing,ikemoto2018noise,krauss2019recurrence,bonsel2021control,metzner2022dynamics}. Additionally, RNNs have been proposed as powerful tools for modeling neural processes in neuroscience \cite{barak2017recurrent}. In particular sparse RNNs, which resemble the human brain in their relatively low average node degree \cite{song2005highly}, have been shown to possess remarkable properties, including enhanced information storage capabilities \cite{brunel2016cortical,narang2017exploring,gerum2020sparsity,folli2018effect}.

\NI In our own previous research, we have systematically analyzed the relation between network structure and dynamical properties, starting with small recurrent three-neuron motifs \cite{krauss2019analysis}. We then demonstrated how statistical parameters of the weight matrix can be used to control the dynamics in large, freely running RNNs \cite{krauss2019weight, metzner2022dynamics}. Another major focus of our past work have been various noise-induced resonance phenomena in these complex dynamical systems \cite{bonsel2021control, schilling2022intrinsic, krauss2016stochastic, 
krauss2019recurrence,
schilling2021stochastic, schilling2023predictive,
metzner2024recurrence}. 

\NI As a next step, we now move from free-running RNNs to the field of neural reservoir computing: Untrained, randomly connected RNNs, so-called reservoirs, can be used to perform actual tasks, in combination with a simple readout layer that is optimized 'in one shot' by using highly effective techniques from linear algebra. In this still unconventional type of neural data processing, some of the reservoir neurons receive external input signals, often in a sequential manner, which are then incorporated into the ongoing recurrent dynamics of the system, where they are gradually spreading over the entire network and becoming nonlinearly transformed. At the end of each input episode, the momentary state of the reservoir represents a high-dimensional and (randomly) processed representation of the sequential input data. The role of the readout layer is then to 'pick out' from the reservoir state all information that is useful in the context of solving a predefined task.

\NI Reservoir computing includes diverse models such as liquid state machines \cite{maass2002real}, echo state networks \cite{jaeger2001echo}, extreme learning machines \cite{huang2015extreme}, and the concept of winnerless competition \cite{rabinovich2000dynamical}. In this work, we are using deterministic RNNs, consisting of neurons with $\tanh$-activation functions that are connected randomly. The magnitudes of the connections are drawn from Gaussian distributions, and the overall structure of the weight matrix is controlled by certain statistical parameters, such as the general strength $w$ of the connections (the standard deviation of the normal weight distribution), the density $d$ of non-zero connections, and the balance $b$ between excitatory and inhibitory connections. Using insights from our former work \cite{krauss2019weight, metzner2022dynamics}, we can thus control the dynamical state of the RNN by these statistical parameters, in particular by $w$ and $b$. 

\NI Here, we are interested in the relationship between a reservoir's dynamical state and its performance in a specific classification task. For this purpose, the actual reservoir dynamics is measured by the fluctuations and correlations of neural activations over time, as well as the degree of nonlinearity in the neural transformations. Together, these quantities allow us to asses the system's dynamical attractor state. Simultaneously, the classification performance is quantified by the accuracy, that is, the fraction of correctly predicted labels with respect to ground truth. 

\NI The optimal dynamical regime of RNNs for computational purposes has been a question of considerable interest over the past decades. For a long time, the dominating opinion was that the 'edge of chaos' \cite{langton1990computation,jaeger2001echo,maass2002real}, a border regime between over-sensitive chaotic dynamics and rigid oscillatory or fixpoint behavior, would probably be the 'sweet spot' for reservoir computing and for similar techniques that exploit an untrained complex dynamical system for data processing. However, more recent work has challenged this hypothesis. For example, Carroll \cite{carroll2020reservoir} tested different reservoir computer designs and discovered that while some configurations did indeed perform optimally near the edge of chaos, others performed better away from this point.

\NI Using our $\tanh$-reservoir, followed by a linear readout layer that is 'trained' for each classification task with the pseudoinverse technique, as well as a subsequent argmax-function to produce definite predicted class labels, we come to a slightly more nuanced conclusion: The edge of chaos is, after all, a preferred dynamical regime for computational RNNs, at least for our investigated tasks and in a certain range of recurrent coupling strengths $w$. For relatively large couplings $w$, as we continuously tune the excitatory/inhibitory balance $b$ from the extreme value -1 (where only inhibitory connections are present, generating an oscillatory reservoir dynamics) to +1 (where only excitatory connections are present, thus trapping the reservoir in a global fixpoint), a plot of the accuracy $A(b)$ as a function of balance $b$ shows indeed two 'edge-of-chaos' peaks, one between the oscillatory and chaotic regimes, and another between the chaotic and fixpoint regime. On the other hand, we find that the 'data processing power' of the reservoir remains intact also for rather small coupling strengths $w$, while the spontaneous dynamics of the reservoir is greatly suppressed in this weak coupling regime. Surprisingly, this makes the accuracy $A(b)$ almost independent from the balance $b$ - a very robust feature of RNNs that might also be exploited in biological neural networks.

\section{Methods}

\subsection {Design of Reservoir Computer (RC)}

Our RC consists of an input-layer, the reservoir, and a readout-layer (compare Fig.\ref{fig_RCAndFreeDyn}). The externally provided input data consists of $E$ subsequent episodes (with each episode corresponding, for example, to a pattern that is supposed to be recognized). This data stream is continuously circulating within the reservoir and simultaneously moving through the system to the output, as described in the following section: 

\NI In each time step $t$, the input layer receives $M$ external data signals $x_m^{(t)}\in\left[-1,+1\right]$ in parallel. These signals are fed into an input matrix $\mathbf{I}$ of size $N\!\times\!M$, which transforms them linearly and then feeds them into the reservoir according to Eq.\ref{yEq}. A complete input episode lasts $T$ time steps, and only after each episode is complete the state of the reservoir is used to predict the next output of the RC. The total number of input signals in each episode is therefore $M\!\cdot\!T$.

\NI The {\bf input layer} of the RC consists only of a matrix $\mathbf{I}$ and is thus purely linear. For simplicity, we set the elements of the input matrix to $I_{mm}\!=\!w$ for $m=0\ldots M\!-\!1$, and $I_{mn}\!=\!0$ otherwise. This means that only the first $M$ of the reservoir neurons receive the direct input signals $x_m^{(t)}$, each scaled by the same factor $w$ (corresponding to the recurrent coupling strength parameter, introduced below). 

\NI The {\bf reservoir} contains $N$ neurons with activation functions of the form $y_n = \tanh(u_n)$, which are all updated simultaneously in each time step. For every neuron $n$, the total momentary input $u_n$ includes a constant bias term $b_{w,n}$, a weighted sum of input signals, and a weighted sum of recurrent neural activations, according to Eq.\ref{yEq}. The initial states $y_n^{(0)}$ of the neurons are assigned to random numbers, distributed uniformly within $\left[-1,+1\right]$. The same initial states are used for repeated simulation runs of the same reservoir. However, reservoirs with different recurrent weights also get different initial states.

\NI The {\bf readout layer} consists of an affine-linear transformation of the reservoir neuron's activations $y_n$, described by a $K\!\times\!N$ output matrix $\mathbf{O}$ and a output bias vector $\mathbf{b_o}$, according to Eq.\ref{zEq}. The weights and biases of this affine-linear transformation are optimized for the desired task by using the method of the pseudoinverse (see below). As a result of this 'learned' transformation, we obtain $K$ output signals $z_k$. Since we consider only classification tasks in this work, each output signal is used to 'vote for' one of the $K$ data classes, ideally in a 'one-hot' manner. In order to obtain a discrete predicted class label $c$ after each input episode, we finally apply the argmax function to the output signals $z_k$. By this way, the readout layer also contains a nonlinear processing step, which helps to create sharp class boundaries in the data space.

\NI Summing up, the RC is described by the following coupled equations:

\begin{eqnarray}
y_n^{(t)} & = & \tanh \left( 
  b_{w,n} + \sum_m I_{nm} \; x_m^{(t-1)} + 
  \sum_{n^{\prime}} W_{nn^{\prime}} \; y_{n^{\prime}}^{(t-1)} 
\right) \label{yEq} \\
z_k^{(t)} & = & b_{o,k} + \sum_n O_{kn} \; y_n^{(t)} \label{zEq} \\
c^{(t)} & = &\operatorname*{arg\,max} \left\{ z_k^{(t)} \right\}
\end{eqnarray}


\subsection{Assignment of reservoir connection weights and biases}

The weight matrix $\mathbf{W}$ of the reservoir's recurrent connections is random, but controlled by three statistical parameters, namely the {\bf density} $d$ of non-zero connections, the excitatory/inhibitory {\bf balance} $b$, and the recurrent {\bf coupling} strength $w$, given by the width of the Gaussian distribution of weight magnitudes. The density ranges from $d=0$ (isolated neurons) to $d=1$ (fully connected network). The balance ranges from $b=-1$ (purely inhibitory connections) to $b=+1$ (purely excitatory connections), with $b=0$ corresponding to a perfectly balanced system. The coupling strength $w$ can take any positive value, with $w=0$ corresponding to unconnected reservoir neurons.

\NI In order to construct a $N\!\times\!N$ weight matrix with given parameters $(b,d,w)$, we first generate a matrix $M^{(magn)}$ of weight magnitudes, by drawing the $N^2$ matrix elements independently from a zero-mean normal distribution with standard deviation $w$ and then taking the absolute value. Next, we generate a random binary matrix $B^{(nonz)} \in \{0,1\}^{N\!\times\!N}$, where the probability of a matrix element being $1$ is given by the density parameter $d$, that is, $p_1 = d$. Next, we generate another random binary matrix $B^{(sign)} \in \{-1,+1\}^{N\!\times\!N}$, where the probability of a matrix element being $+1$ is given by $p_{+1} = (1+b)/2$, where $b$ is the balance parameter. Finally, the weight matrix is constructed by element-wise multiplication, that is, $W_{mn}=M^{(magn)}_{mn}\cdot B^{(nonz)}_{mn}\cdot B^{(sign)}_{mn}$. Note that throughout this paper, the density parameter is always set to the maximum of $d=1$.

\NI The biases $b_{w,n}$ of the reservoir are drawn independently from a zero-mean normal distribution with standard deviation $w^{\prime}$. Note that throughout this paper, the strength of the reservoir bias is always set to $w^{\prime}=0.1$. Thus, even for an uncoupled reservoir with $w=0$ and with zero input, the neurons will have non-zero resting values according to the random biases.


\subsection{Optimal readout layer using pseudoinverse}

\NI The weights and biases of the readout layer are computed using the pseudoinverse in the following way: 

\NI Let \( Y \in \mathbb{R}^{(E-1) \times N} \) be the matrix of reservoir states directly after each input episode, where \( E \) is the total number of episodes and \( N \) is the number of reservoir neurons. Let \( Z \in \mathbb{R}^{(E-1) \times K} \) be the matrix of target output states, where \( K \) is the number of output units.

\NI To account for biases in the readout layer, a column of ones is appended to \( Y \), resulting in the matrix \( Y_{\text{bias}} \in \mathbb{R}^{(E-1) \times (N+1)} \):

\[
Y_{\text{bias}} = \begin{bmatrix} Y & \mathbf{1}_{E-1} \end{bmatrix}
\]

\NI where \( \mathbf{1}_{E-1} \in \mathbb{R}^{(E-1) \times 1} \) is a column vector of ones.

\NI The weights and biases of the readout layer are computed by solving the following equation using the pseudoinverse of \( Y_{\text{bias}} \):

\begin{equation}
W_{\text{bias}} = Y_{\text{bias}}^+ Z
\label{optWei}
\end{equation}

\NI where \( Y_{\text{bias}}^+ \) is the Moore-Penrose pseudoinverse of \( Y_{\text{bias}} \), and \( W_{\text{bias}} \in \mathbb{R}^{(N+1) \times K} \) contains both the readout weights and the biases.

\NI To compute the pseudoinverse, we first perform a singular value decomposition (SVD) of \( Y_{\text{bias}} \):

\[
Y_{\text{bias}} = U S V^\top
\]

\NI where \( U \in \mathbb{R}^{(E-1) \times (E-1)} \) is a unitary matrix, \( S \in \mathbb{R}^{(E-1) \times (N+1)} \) is a diagonal matrix containing the singular values, and \( V^\top \in \mathbb{R}^{(N+1) \times (N+1)} \) is the transpose of a unitary matrix.

\NI The pseudoinverse of \( Y_{\text{bias}} \) is computed as:

\[
Y_{\text{bias}}^+ = V S^+ U^\top
\]

\NI where \( S^+ \in \mathbb{R}^{(N+1) \times (E-1)} \) is the pseudoinverse of the diagonal matrix \( S \). The pseudoinverse \( S^+ \) is obtained by taking the reciprocal of all non-zero singular values in \( S \) and leaving zeros unchanged.

\NI Finally, after inserting  \( Y_{\text{bias}}^+ \) into Eq.\ref{optWei}, the optimal readout weights \( W \in \mathbb{R}^{K \times N} \) and biases \( b \in \mathbb{R}^K \) are extracted from the extended matrix \( W_{\text{bias}} \) as

\[
W = W_{\text{bias}}^\top[:N]
\]
\[
b_w = W_{\text{bias}}^\top[N]
\]


\subsection {Fluctuation measure}

The degree $F$ of neural fluctuations in the reservoir is determined by first computing the standard deviations $\sigma_n$ of the temporal activations $y_n^{(t)}$ during the entire simulation period, individually for each neuron $n$. Then $F=\left\langle \sigma_n\right\rangle_n$ is defined as the mean of these standard deviations. Since the $\tanh$-neurons are producing output values $y$ in the range $\left[-1,+1\right]$, the measure $F$ is always in the range $\left[0,1\right]$. The value $F\!=\!0$ is characteristic for a resting reservoir, or one in a fixpoint state. The value $F\!=\!1$ occurs in the oscillating state of the reservoir, where each neuron flips between activations -1 and +1.


\subsection {Correlation measure}

The degree $C$ of correlations between subsequent activation states of the reservoir neurons is determined by first computing the linear correlation matrix $C_{mn}=\left\langle y_m^{(t)}\cdot y_n^{(t\!+\!1)} \right\rangle_t$ for each pair $(m,n)$ of neurons. Then $C=\left\langle C_{mn} \right\rangle_{mn}$ is defined as the mean of these correlation matrix elements. Since the $\tanh$-neurons are producing output values $y$ in the range $\left[-1,+1\right]$, the measure $C$ is always in the range $\left[-1,+1\right]$. The value $C\!=\!0$ is characteristic for a resting reservoir, or one in a highly chaotic state. The value $C\!=\!-1$ occurs in the oscillating state of the reservoir. The value $C\!=\!+1$ occurs in a fixpoint state.


\subsection {Nonlinearity measure}

\NI The $\tanh$-neurons are producing output values $y$ in the range $\left[-1,+1\right]$, and by pooling over all neurons and time steps of a run we can generate a distribution $p(y)$ of neural activations.

\NI If $p(y)$ has a single central peak at $y\!=\!0$, the system is in the linear working regime, whereas a distribution with two peaks close to $y\!=\!-1$ and $y\!=\!+1$ corresponds to the nonlinear regime.

\NI To quantify the degree of nonlinearity, we introduce the measure $\alpha=f_A-f_B+f_C$, where $f_A$ is the fraction of $y$-values in the interval $\left[-1,-0.5\right[$, $f_B$ the fraction in $\left[-0.5,+0.5\right]$, and $f_C$  the fraction in $\left]+0.5,+1\right]$. 

\NI The measure $\alpha$ is always in the range $\left[-1,+1\right]$. One obtains $\alpha\!\approx\!-1$ in the strongly linear (analog) regime, $\alpha\!\approx\!0$ in the intermediate regime when the distribution of activations $p(y)$ is flat, and $\alpha\!\approx\!+1$ in the strongly nonlinear (digital) working regime of the reservoir.


\subsection {Accuracy measure}

In this work, the RC is trained and tested for various model classification tasks. Since our data distributions are artificially generated, we can 'draw' an arbitrary number of input patterns from such a distribution, each consisting of $M\!\cdot\!T$ real values that are presented during an episode, and each belonging to one of $K$ non-overlapping classes.

\NI After the readout-layer has been fitted to the training data set, the RC is successively presented $E$ episodes of an independent test data set and predicts a series of $E$ discrete class labels $c_0,c_1,\ldots,c_{E\!-\!1}$, with $c_i\in\left\{0\ldots K\!-\!1\right\}$. They are compared to the ground truth labels $l_0,l_1,\ldots,l_{E\!-\!1}$, and the accuracy $A\in\left[0,1\right]$ is defined as the fraction of correct predictions, where $c_i\!=\!l_i$. Since the $K$ classes are presented with equal frequencies in the input data sets, the chance level accuracy is $A_{chance}=\frac{1}{K}$ in our case.


\subsection{List of symbols}

\begin{table}[ht!]
  \centering
  \begin{tabular}{|c|p{10cm}|}
    \hline
    \textbf{Symbol} & \textbf{Meaning} \\ 
    \hline
    \hline
    $M$ & Nr. of input channels \\
    $N$ & Nr. of reservoir neurons \\
    $K$ & Nr. of output channels/classes \\
    $T$ & Nr. of time steps per input episode \\
    $E$ & Nr. of episodes in data set \\
    $R$ & Nr. of independent simuluation runs \\
    \hline
    $\mathbf{I}$ & input weight matrix \\
    $\mathbf{W},\mathbf{b}_w$ & reservoir weight matrix and bias \\
    $\mathbf{O},\mathbf{b}_o$ & output weight matrix and bias \\
    \hline
    $x$ & input signals \\
    $y$ & reservoir neural activations \\
    $z$ & contin. output signals before argmax \\
    $c$ & discrete predicted class label \\
    \hline
    $w$ & strength of recurrent coupling \\
    $w^{\prime}$ & strength of reservoir bias \\
    $d$ & density of non-zero connections \\
    $b$ & excitatory/inhibitory balance \\
    \hline
    $F$ & fluctuation of neural activations \\
    $C$ & correlation of neural activations \\
    $\alpha$ & nonlinearity of neural activations \\
    $A$ & accuracy of predicted labels \\
    \hline
  \end{tabular}
\end{table}

\section{Results}

A reservoir computer consists of an input-layer, the reservoir, and a readout-layer (for details, see Fig.\ref{fig_RCAndFreeDyn}(a) and the Methods section). The reservoir, a nonlinear recurrent neural network (RNN), is used for a variety of computational functions, such as storing past input information, dimensionality expansion, and nonlinear data transformation. However, these RNNs are complex dynamical systems, and therefore they often show surprising emergent effects, even in isolation from the input and readout layer. Moreover, the readout layer, together with the optimization technique of the pseudoinverse, is a powerful learning engine by its own, able to exploit very subtle changes in the reservoir dynamics for the task at hand. In order to understand the behavior of the whole reservoir computer, it is therefore essential to first analyze its individual parts and only then consider their complex interplay. 

\subsection{Reservoir dynamics without input}

We start with an investigation of an isolated, free-running reservoir of $N\!=\!10$ neurons with $\tanh$ activation functions. The neurons are mutually connected according to a reservoir weight matrix $\mathbf{W}$ that is random, but controlled by two statistical parameters: The overall strength $w$ of the recurrent coupling and the balance $b$ between excitatory and inhibitory connections (See Methods section for details). 

\NI By changing these two control parameters, we can tune the reservoir into various dynamical regimes, which are marked by characteristic global behaviors of the neural activations, such as oscillations, chaotic fluctuations, or trapping in fixed points of state space. 

\NI These different types of dynamical behaviors, throughout this paper, are quantified by three measures: The degree $F$ of fluctuation in the neural activations, the correlation $C$ between subsequent activations, and finally the degree of nonlinearity $\alpha$, which reflects the shape of the distribution $p(y)$ of neural activations (See Methods section for details). 

\begin{figure}[ht!]
\centering
\includegraphics[width=1\linewidth]{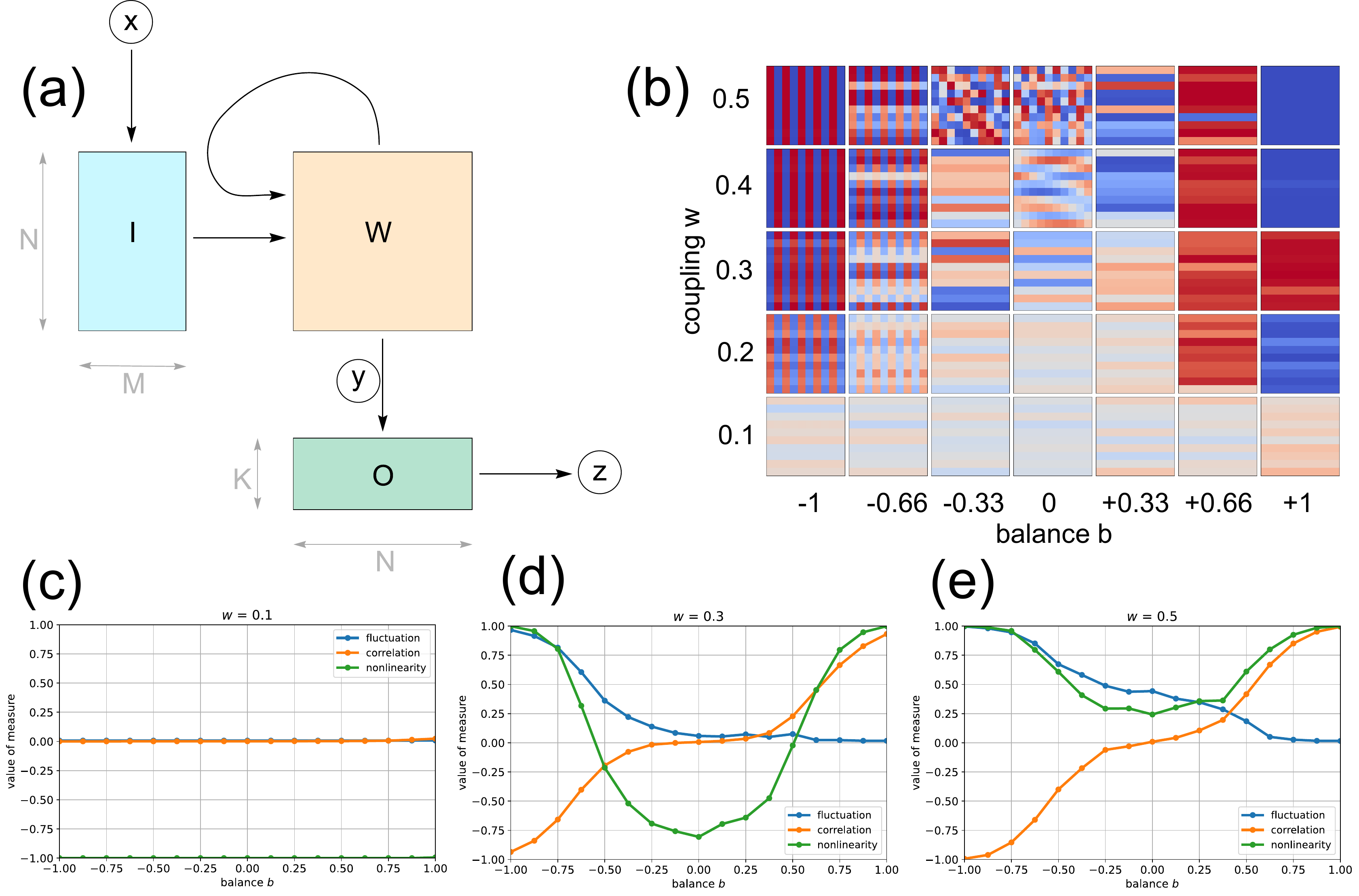}
\caption{{\bf Reservoir computer and free reservoir dynamics.}
{\bf (a)} In every time step $t$ of an ongoing computation, an input matrix $\mathbf{I}$ of size $N\!\times\!M$ couples $M$ input signals $x_m$ into the reservoir. The $N$ neurons of the reservoir, which are recurrently connected by a matrix $\mathbf{W}$ of size $N\!\times\!N$, produce neural activations $y_n$. An output matrix $\mathbf{O}$ of size $K\!\times\!N$ reads the reservoir states $y_n$ and linearly extracts from them $K$ output features $z_k$. These features are finally passed through a nonlinear {\em argmax} function (not shown) to produce a one-hot output for classification tasks.
{\bf (b)} Examples for the dynamics in a free-running reservoir with $N\!=\!10$ neurons, for seven different values $b$ of the excitatory/inhibitory balance and five different recurrent coupling strengths $w$ in the reservoir's connection matrix $\mathbf{W}$. For each parameter combination $(w,b)$ we show the momentary activations $y_n^{(t)}\in\left[ -1,+1\right]$ of the 10 neurons (vertical) over time (horizontal) in color coding, where blue indicates negative and red positive activations (Comp. Fig.\ref{fig_ModelTasks}(g)).
{\bf (c),(d),(e)}  Three measures of reservoir dynamics as a function of the excitatory/inhibitory balance $b$, for three different recurrent coupling strengths $w=0.1$ (left), $w=0.3$ (middle) and $w=0.5$ (right). The {\bf fluctuation} $F\in\left[0,1\right]$ measures how much, averaged over all 10 reservoir neurons, the individual neural activations $y_n^{(t)}$ change over time. It usually has the largest values in the oscillatory regime, medium values in the chaotic regime, and the smallest values in the fixpoint regime (blue lines). The {\bf correlation} $C\in\left[-1,+1\right]$ measures the average degree of similarity between the neural activation $y_i^{(t)}$  of neuron $i$ at time $t$ and the activation $y_j^{(t\!+\!1)}$  of neuron $j$ at the following time step $t\!+\!1$, averaged over all pairs $(i,j)$. It is usually negative in the oscillatory regime, around zero in the chaotic regime, and positive in the fixpoint regime (orange lines). The {\bf nonlinearity parameter} $\alpha\in\left[-1,+1\right]$ measures how strongly the reservoir neurons are driven into the saturation of the $\tanh$ activation functions and therefore produce 'digital' (-1 or +1) instead of 'analog' (continuous) outputs. The parameter $\alpha$ is close to -1 (linear, analog regime) when the distribution $p(y)$ of activations has a single peak at $y\!\approx\!0$, it is close to 0 if the distribution is distributed uniformly in the range $\left[-1,+1\right]$ (weakly nonlinear regime), and it is close to +1 (nonlinear, digital regime) if $p(y)$ has its peaks at the borders $y\!\approx\!-1$ and/or $y\!\approx\!+1$ (green lines).
} 
\label{fig_RCAndFreeDyn}
\end{figure}

\NI In the following, while keeping the coupling parameter $w$ at selected fixed values, we scan the balance parameter $b$ between its extreme values -1 and +1. For each combination $(w,b)$ of control parameters, we run simulations for an ensemble of $R\!=\!100$ independent reservoirs, with different weight matrices and starting states. After every run of a reservoir, we compute the three measures $(F,C,\alpha)$, and then average them over the ensemble of (statistically equivalent) reservoirs. Finally, we plot the average measures as functions of the control parameters Fig.\ref{fig_RCAndFreeDyn}(c-e). In addition, we present selected examples of detailed time-dependent reservoir activations in Fig.\ref{fig_RCAndFreeDyn}(b). 

\NI For weak recurrent coupling $w\!=\!0.1$, there is no spontaneous temporal dynamics in the reservoir, and all neural activations $y_n^{(t)}$ rest at the small constant values that are determined by the random bias vector of the reservoir (Fig.\ref{fig_RCAndFreeDyn}(b), lowest row). Since the measure $F$ first computes the individual temporal standard deviation $\sigma_n$ of each neuron's activation and then averages the $\sigma_n$ over all neurons, we obtain $F\!\approx\!0$ in this case (Fig.\ref{fig_RCAndFreeDyn}(c), blue curve and dots). The correlation $C$, which is an average over all possible pairwise time-averaged products of activations, $C_{mn}=\left\langle y_m^{(t)}\cdot y_n^{(t\!+\!1)} \right\rangle_t$, is also close to zero in this regime (orange curve and dots), because the $C_{mn}$ are distributed randomly and symmetrically around zero. The nonlinearity parameter has here the value $\alpha\!=\!-1$, which indicates that all neural activations are small and distributed like a single peak around the center $y\!=\!0$ of the activation range. This corresponds to a linear regime of the reservoir.

\NI When the recurrent coupling is slightly increased to $w\!=\!0.2$, reservoirs with balances around $b\!\approx\!0$ (say, from -0.33 to +0.33) still tend to rest at small constant activation values (Fig.\ref{fig_RCAndFreeDyn}(b), second row from below). However, as a first coupling-induced 'phase transition' of the reservoir, systems with balances close to $b\!\approx\!-1$ now start to fall into a globally oscillating attractor (with synchronous flipping between activations -1 and +1), whereas reservoirs with balances close to $b\!\approx\!+1$ fall into a global fixed point (where all neurons rest either at constant activation +1 or constant activation -1).

\NI In the case $w\!=\!0.3$ of medium coupling strength (Fig.\ref{fig_RCAndFreeDyn}(d)), the oscillatory and fixed point behaviors at $b\!\approx\!-1$ and $b\!\approx\!+1$, respectively, are again apparent. In the globally oscillating attractor (left side of scan plot), the fluctuation approaches $F\!\approx\!1$ (blue) and the correlation $C\!\approx\!-1$ (orange). By contrast, in the fixed point attractor (right side of scan plot), the fluctuation is $F\!\approx\!0$ and the correlation is $C\!\approx\!+1$. The degree of oscillatory or fixed point behavior, quantified by $F$ and $C$, is continuously decreasing as we move from the extreme balances $b\!=\!-1$ or $b\!=\!+1$ toward $b\!=\!0$. In the perfectly balanced system at $b\!=\!0$, we still find $F\!\approx\!0$ and $C\!\approx\!0$, indicating temporally constant neural activations. However, now the nonlinearity parameter has increased to $\alpha\!\approx\!-0.75$, which means that the distribution of (still time-independent) activations has widened, compared to smaller couplings $w$. This is confirmed by the central square in Fig.\ref{fig_RCAndFreeDyn}(b).

\NI In the case $w\!=\!0.5$ of strong coupling (Fig.\ref{fig_RCAndFreeDyn}(e)), we are already beyond a second 'phase transition' of the reservoir, as the balanced system ($b\!=\!0$) now shows chaotic (uncorrelated) spontaneous fluctuations, indicated by $C\!\approx\!0$ but $F\!>\!0$. The chaotic behavior also extends to slightly unbalanced systems. However, for the extreme values $b\!=\!-1$ or $b\!=\!+1$, we still find oscillatory or fixed point attractors. The nonlinearity parameter $\alpha$ has now become positive for all balances, indicating that the distribution $p(y)$ of neural activations has switched from a peak at $y=0$ to peaks at $y=-1$ and/or $y=+1$. This corresponds to a more 'digital' regime, where the neurons are frequently driven into the saturation of their activation function. In the top row of Fig.\ref{fig_RCAndFreeDyn}(b), this is indicated by the appearance of stronger colors over the whole range of balances, as compared to the bottom row.

\subsection{Model classification tasks}

\begin{figure}[ht!]
\centering
\includegraphics[width=0.8\linewidth]{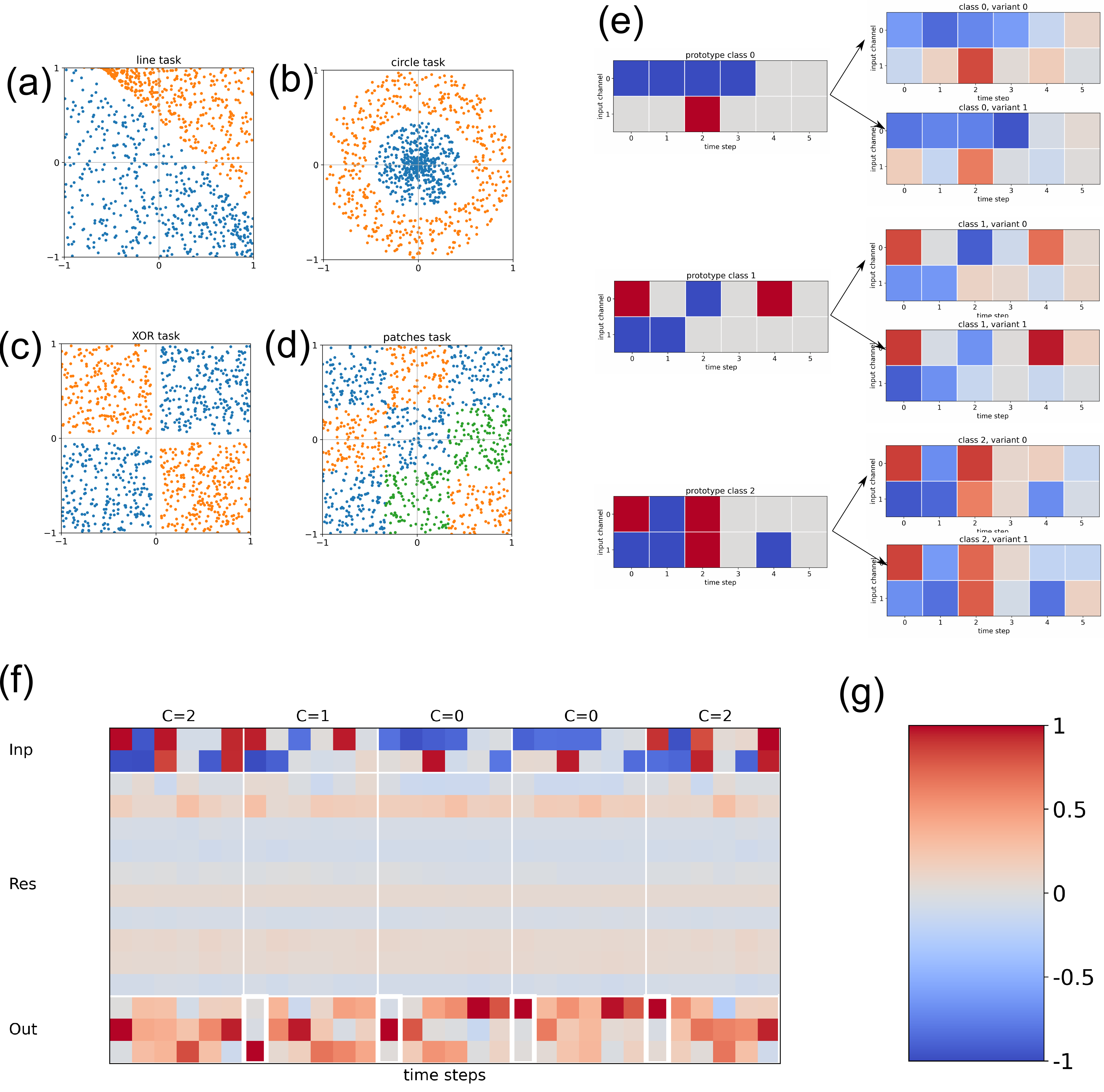}
\caption{{\bf Model classification tasks and states of the running RC.} The 'purely spatial' discrimination tasks (a-d) have only $M\!=\!2$ simultaneous input signals $x_0\in\left[-1,+1\right]$ and $x_1\in\left[-1,+1\right]$ per episode, and thus each input vector $(x_0,x_1)$ can be represented as a point in a two-dimensional plane, with point colors representing the classes. By contrast, task (e) is a 'spatio-temporal' pattern recognition task.
{\bf (a)}: 'Line task': Two point classes that can be separated by a line in input space.  
{\bf (b)}: 'Circle task': Two point classes that can be separated by a circle. 
{\bf (c)}: 'XOR task': Two point classes that cannot be separated by any single curve.
{\bf (d)}: 'Patches task': Three point classes, each distributed over several patches in input space.
{\bf (e)}: A task where, in each episode, $M\!=\!2$ simultaneous input signals are subsequently presented over $T\!=\!6$ time steps. The task requires to discriminate $C\!=\!3$ classes, and each is represented by a prototypical pattern of $M\!\cdot\!T$ ternary values $x_m^{(t)}\in\left\{-1,0,+1\right\}$ (color-coded by blue, gray and red). From each prototypical pattern, variants can be produced by random gradual modification of its characteristic values (Only two variants per prototype are shown). These modifications are done in a way that preserves the range $x_m^{(t)}\in\left[-1,+1\right]$ of input signals. Note that in task (e), the final input values are zero (gray) in each of the prototypical patterns, so that the readout-layer cannot simply use the terminal pair of input-values to discriminate the classes. The number of 'terminal zeros' in each pattern can be used to test the memory capacity of the reservoir.
{\bf (f)}: Detailed time-dependent states of the RC during 5 consecutive input episodes. The horizontal axis represents the time steps, the vertical axis represents the input signals $x$ (top 2 rows), the activations $y$ of the reservoir neurons (middle 10 rows), and the linear outputs $z$ before argmax (bottom 3 rows). Episode boundaries are marked by vertical white lines, and the correct class labels of each input episode is written on top. Predicted labels appear in one-hot coding directly after each episode end (framed in white).
{\bf (g)}: General color bar for matrix plots in this paper.
}
\label{fig_ModelTasks}
\end{figure}

Having analyzed the dynamical properties of the isolated reservoir, we will next feed into the RC continuous input signals, which are always kept in the standard range $\left[-1,+1\right]$. As described in the Method section, in every time step $t$, the first $M$ reservoir neurons receive a new input vector $\mathbf{x}^{(t)}$, consisting of $M$ simultaneously applied input signals $x_m^{(t)}$, scaled with a factor $w$ by the input matrix. The ongoing inputs are presented in the form of subsequent 'episodes', each with a temporal length of $T$ time steps, and the $M\!\cdot\!T$ input values of an episode can be viewed as one 'data packet' that is to be processed by the RC. The readout-layer will compute its output signals based on the momentary state of the reservoir at the time step directly after each episode (that is, when the first vector $\mathbf{x}^{(T\!+\!1)}$ of the next input episode is already fed in. Compare Fig.\ref{fig_ModelTasks}(f)).

\NI In this work, we will not use 'real-world' tasks for the RC, but artificial classification tasks that can be controlled in their complexity, and for which data sets of arbitrary size can be generated. The goal of the RC will be to predict a discrete class label $c\in\left\{0,\ldots,C\!-\!1\right\}$ for each input data packet. 

\NI In a first type of 'purely spatial' tasks (Fig.\ref{fig_ModelTasks}(a-d)), the $M$ parallel input signals remain constant during the whole episode, so that the RC does not really need to memorize a temporal sequence. In the second, 'spatio-temporal' type (Fig.\ref{fig_ModelTasks}(e)), the input signals change during the episode. For such tasks, the demands on the memory capacity of the reservoir can be further intensified by ending each episode with a certain number of zero input vectors.  

\subsection{Reservoir dynamics with input}

We next investigate how the continuous feed-in of input signals influences the dynamics of the reservoir in its different regimes. For these numerical experiments, we use again 10-neuron reservoirs (like in Fig.\ref{fig_RCAndFreeDyn}(b-e)) and continuous inputs from a spatio-temporal classification task (similar to Fig.\ref{fig_ModelTasks}(e,f)). This type of task has the strongest temporal fluctuation in the input signals.

\begin{figure}[ht!]
\centering
\includegraphics[width=1\linewidth]{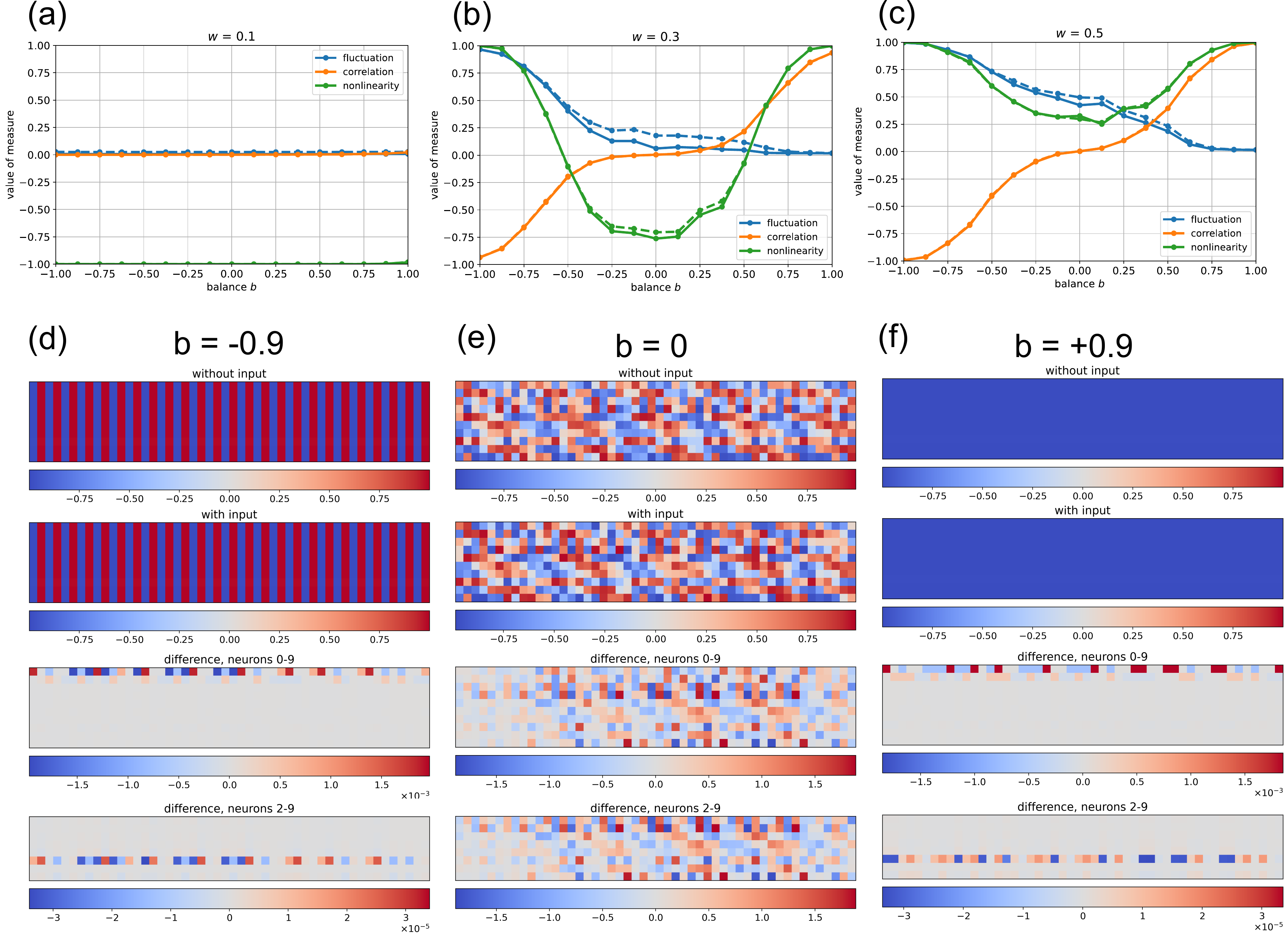}
\caption{{\bf Reservoir dynamics with input.}
The top panels (a-c) show the three quantitative statistical measures $F$ (blue), $C$ (orange) and $\alpha$ (green) of the reservoir dynamics as functions of the balance $b$, for three different recurrent couplings $w$, both without external input signals (solid lines) and with continuously fed-in spatio-temporal input signals (dashed lines).
{\bf (a)}: For weak coupling $w\!=\!0.1$, there is no significant statistical difference between the free-running and input-driven reservoir.
{\bf (b)}: For medium coupling $w\!=\!0.3$, the correlation is not affected by the input, but the fluctuation and nonlinearity are slightly enhanced by the input around the chaotic regime.
{\bf (c)}: For strong coupling $w\!=\!0.5$, only the fluctuation is slightly increased by the input.
The bottom panels (d-f) show the time-dependent neuron activations for the strongly coupled reservoir with $w\!=\!0.5$, for three different values of the balance $b$. First row: activations without input. Second row: activations at the same times, but with spatio-temporal input signals. Third row: difference between the input-driven and free-running activations, for all reservoir neurons. Fourth row: difference for the neurons 2-9 that do not receive direct input signals.
{\bf (d)}: In the oscillatory regime $b=-0.9$, the input causes activation differences in the neurons 2-9 that are only of order $10^{-5}$, compared to the activations themselves, which are of order unity.
{\bf (e)}: In the chaotic regime $b=0$, the input-induced differences are considerable and of the same order than the activations themselves.
{\bf (f)}: In the fixpoint regime $b=+0.9$, just like in the oscillatory regime, the input-induced differences are extremely small. Together, this figure demonstrates that with the exception of the sensitive chaotic regime, the input signals only represent a negligible perturbation of the ongoing reservoir dynamics. 
}
\label{fig_ResWithInp}
\end{figure}

\NI We first investigate the three statistical measures of reservoir dynamics as functions of the excitatory/inhibitory balance $b$. In the weakly coupled, free-running reservoir with $w\!=\!0.1$, we again find $F\!=\!0$, $C\!=\!0$ and $N\!=\!-1$ independent from the balance. Interestingly, this 'calm reservoir' condition is not perturbed to any significant degree when input signals are applied (Fig.\ref{fig_ResWithInp}(a)).

\NI We next increase the recurrent coupling $w$ of the reservoir (Fig.\ref{fig_ResWithInp}(b,c)) and again compare the $b$-dependent statistical measures of the free-running (solid lines) and input-driven (dashed lines) systems. Here, we find for medium coupling strength $w\!=\!0.3$ a small but significant input-induced enhancement of the fluctuation $F$ and of the nonlinearity parameter $\alpha$, in particular for balances $b$ around zero, whereas the correlation $C$ seems not affected. Analogous, but even smaller differences are found in the stronger coupled reservoir with $w\!=\!0.5$.

\NI To study the strongly coupled case $w\!=\!0.5$ in more detail, we directly compare the neural activations of the same reservoir under free-running and input-driven conditions (Fig.\ref{fig_ResWithInp}(d-f)). We find that in the oscillatory regime ($b\!=\!-0.9$, (d)), as well as in the fixed point regime ($b\!=\!+0.9$, (f)), the input-induced differences in the neural activations are extremely small. Indeed, for the eight neurons that do not receive direct input signals, the 'perturbation' due to the input is of order $10^{-5}$ compared to the activations themselves. Only close to the chaotic regime, occurring in balanced reservoirs ($b\!=\!0$, (e)), the extreme sensitivity of the dynamics on external influences leads to notable differences between the free-running and input-driven system.

\NI Together, these results demonstrate that with the exception of the sensitive chaotic regime, the input signals only represent a negligible perturbation of the ongoing reservoir dynamics.

\subsection{Readout-layer with input but no reservoir}

The reservoir in a RC is serving multiple purposes, including the encoding of sequentially arriving input information in the form of a complex momentary reservoir state, and the nonlinear processing of the input data. The readout layer (consisting in our case of one linear neuron per output channel and an argmax function for producing discrete class labels) is critically depending on those functions of the reservoir, but it also has data processing abilities by its own. 

\NI In order to explore those abilities, we next investigate a readout layer that is directly connected to the input, without the assistance of a reservoir. For this purpose, we use binary classification tasks in which all input information is presented simultaneously, so that the memory capacity of the reservoir is not needed. 

\NI In particular, we focus on the four types of tasks introduced in Fig.\ref{fig_ModelTasks}(a-d), in which two-dimensional input vectors $\mathbf{x}$ are to be mapped onto one of two class labels $c\in\left\{0,1\right\}$. Accordingly, the readout layer (before the argmax) is given two output channels $z_0,z_1$, which are supposed to encode the two labels in one-hot coding, meaning that $c\!=\!0$ is represented by $z_0\!=\!1,z_1\!=\!0$ and $c\!=\!1$ is represented by $z_0\!=\!0,z_1\!=\!1$. 

\begin{figure}[ht!]
\centering
\includegraphics[width=1\linewidth]{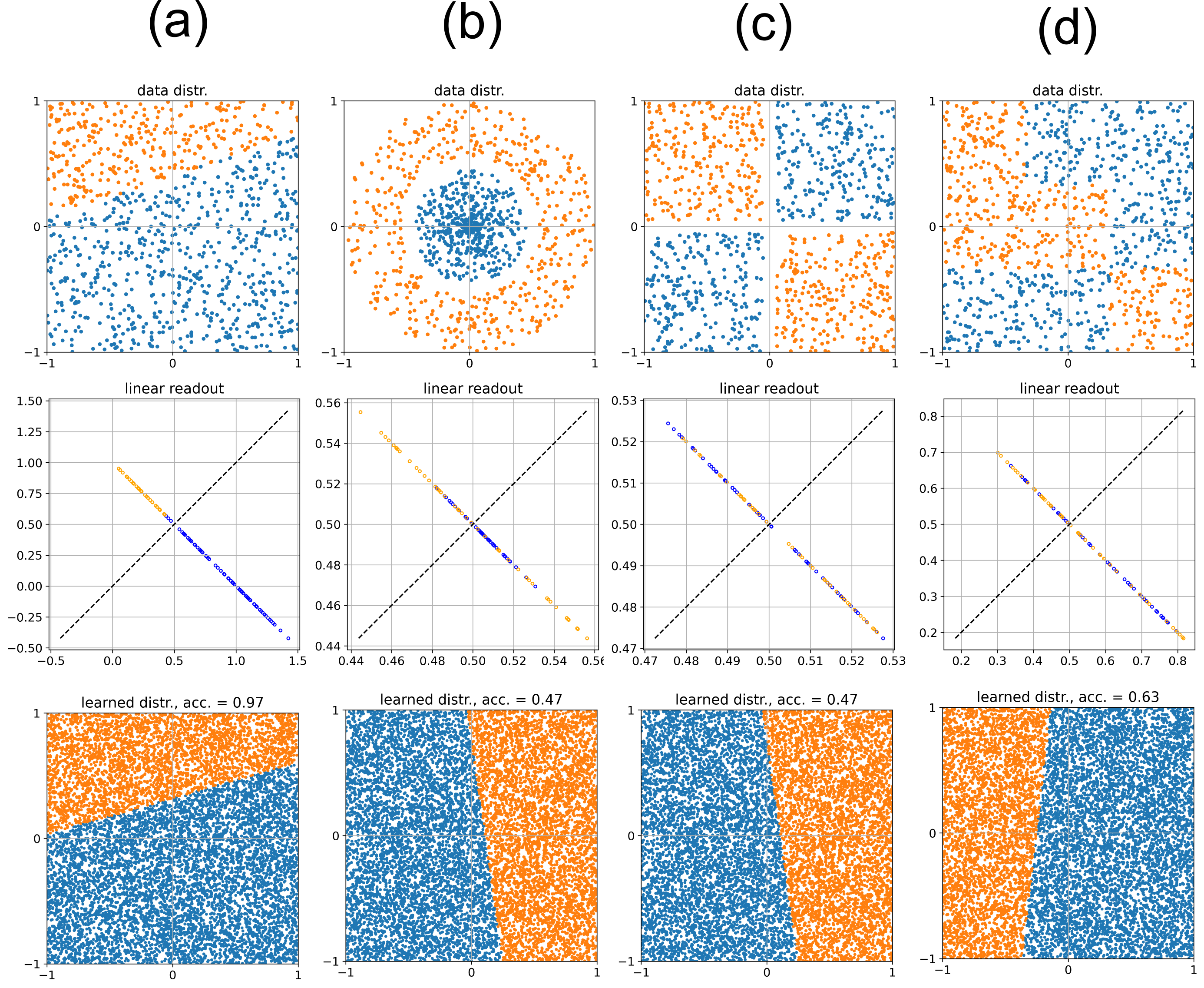}
\caption{{\bf Classification with readout, but no reservoir.} The performance of the isolated readout layer is tested with four purely spatial, binary classification tasks. Top row: input data distributions $p(\mathbf{x})$; Middle row: distribution of the readout layer's linear output $p(\mathbf{z})$, before the application of the argmax function. Due to the training of one-hot outputs, the predictions $\mathbf{z}=(z_0,z_1)$ always lie on a straight line with $z_0\!+\!z_1\!=\!1$. The argmax function separates the points on this line into two discrete classes. The separating boundary is shown as a dashed black line; Bottom row: learned data distributions with achieved accuracy.
{\bf (a)}: The 'line' task is linearly separable. Consequently, even the isolated readout layer achieves a classification accuracy of $0.97$ in this case.
{\bf (b,c)}: The 'circle' and 'XOR' tasks are not linearly separable. Consequently, the accuracy of the isolated readout layer drops to chance level $\approx 0.5$ in these cases.
{\bf (d)}: Although 'patches' tasks are not in general linearly separable, the readout achieves an accuracy slightly above chance level in some cases.
}
\label{fig_ReadoutOnly}
\end{figure}

\NI For each task (Fig.\ref{fig_ReadoutOnly}, first row), a training data set is generated, and the weights and biases of the two readout neurons are computed from the input vectors $\mathbf{x}=(x_0,x_1)$ and the desired output vectors $\mathbf{z}=(z_0,z_1)$ with the method of the pseudoinverse. Then, an independent test data set is generated for the same task and used to determine the classification accuracy of the 'trained' system.

\NI To obtain more insight into the properties of the trained readout layer, we first investigate the distribution of the linear outputs $\mathbf{z}=(z_0,z_1)$, before application of the argmax function (Fig.\ref{fig_ReadoutOnly}, middle row). Due to the prescription of one-hot outputs, all points $\mathbf{z}=(z_0,z_1)$ are positioned on a straight line, according to the condition $z_0\!+\!z_1\!=\!1$. The argmax function then maps each $\mathbf{z}$ on one of two labels, with the separating line given by $z_0\!=\!z_1$ (dashed black lines in Fig.\ref{fig_ReadoutOnly}). We compute and visualize the data distribution actually learned by the system by drawing random points in $\mathbf{x}\in\left[-1,+1\right]^2$ and plotting them with label-specific colors (Fig.\ref{fig_ReadoutOnly}, bottom row).

\NI For the 'line' task (Fig.\ref{fig_ReadoutOnly}(a)), which is obviously linearly separable, the $\mathbf{z}$ for the two classes end up at opposite sides of the separating line (a, middle row), and the resulting accuracy of $0.97$ is close to perfect (a, bottom row). Thus, even an isolated readout layer, without the help of a reservoir, can correctly classify certain types of tasks.

\NI By contrast, the 'circle' and 'XOR' tasks (Fig.\ref{fig_ReadoutOnly}(b,c)) are not linearly separable, and the resulting accuracies of the isolated readout layer are with $0.47$ close to chance level.

\NI The 'patches' tasks (Fig.\ref{fig_ReadoutOnly}(d)), in general, are not separable by a linear line, however some are approximately separable. For example, in the case shown, the readout layer can at least achieve an accuracy of $0.63$ by placing the border between the two half-spaces in data space such that most of the 'blue' class points end up on the same side only (d, bottom row).

\subsection{Neural nonlinearity and RC performance}
\label{cmpLinTanh}

The readout layer needs the assistance of a reservoir to classify data distributions that are not linearly separable. For that purpose, we use in the following numerical experiments a small reservoir with $N\!=\!10$ neurons, weak coupling $w\!=\!0.1$, and excitatory/inhibitory balance $b\!=\!0$. As we have shown before (Fig.\ref{fig_ResWithInp}(a)), despite of the balanced connections, the weakly coupled reservoir remains in a 'calm' state, where the neural activations fluctuate only very slightly (around the small resting values dictated by the biases).  

\NI Investigating the same four (purely spatial) tasks as in the last section, the two-dimensional input vectors $\mathbf{x}$ are now fed into the reservoir during $T\!=\!6$ consecutive time steps in each episode, even though the input does not change during these six time steps. As usual, the readout-layer is using only the reservoir states immediately after each episode to compute valid output.

\NI As a preliminary test, we switch all reservoir neurons from $tanh$ to purely $linear$ activation functions. We then train a fixed reservoir (that is, using in each case the same weights and biases) on the four spatial classification tasks. We find that the resulting linear RC achieves accuracies of $0.98$ in the 'line' task, $0.46$ in the 'circle' task, $0.52$ in the 'XOR' task, and $0.65$ in a 'patches' task. As expected, just as the isolated readout-layer, the linear RC can perfectly handle linear separable problems, to some extend also approximately separable ones, but not cases like the circle or XOR task. In order to classify the circle distribution, the circular boundary between the two classes has to be transformed into a linear boundary, and this is only possible with nonlinear neurons.  

\NI We therefore repeat the above experiments with standard $tanh$ neurons. For the resulting nonlinear RC we obtain accuracies of $0.96$ in the 'line' task, $0.97$ in the 'circle' task, $0.97$ in the 'XOR' task, and $0.93$ in a 'patches' task. In other words, $tanh$ neurons produce the nonlinear data transformations that are required to handle all four classification tasks equally well. 

\NI As a side remark, it is worthwhile to mention that the accuracy of the $\tanh$-RC in the circle task drops to chance level when the random biases of the reservoir neurons are set to zero. As further experiments have shown, this happens because the circle task requires a second-order term in the Taylor expansion of the neuron's activation function around $u\!\approx\!0$. The function $\tanh(u)=a_0u+a_3u^3+\ldots$ does not have such a quadratic term, whereas the biased function $\tanh(u+c)$ does. By contrast, RCs with even activation functions, such as $\exp(-u^2)$ or $\cos(u)$, have a quadratic term and thus can handle the circle problem even without biases.

\begin{figure}[ht!]
\centering
\includegraphics[width=1\linewidth]{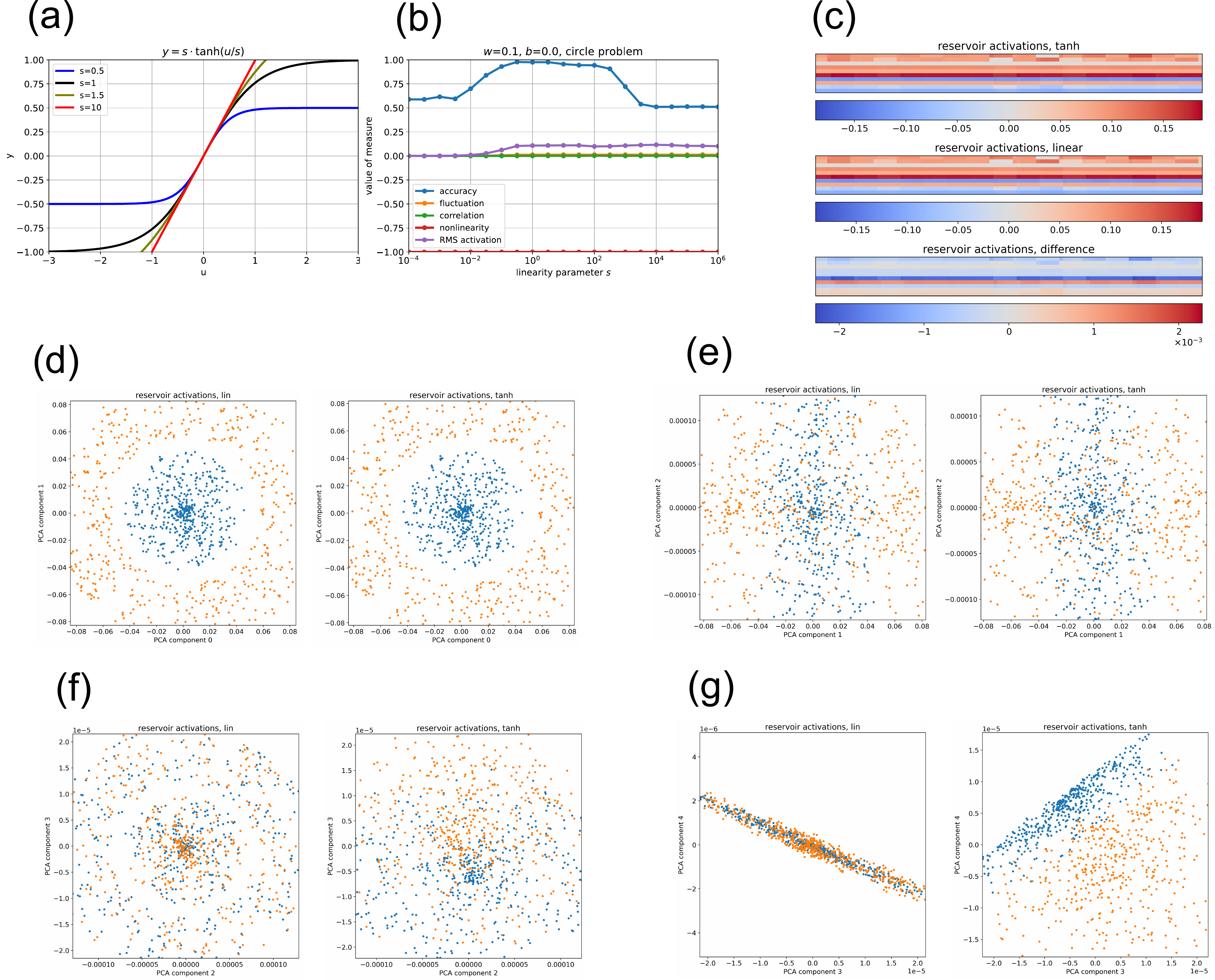}
\caption{{\bf Effect of neural nonlinearity in the RC.} 
{\bf (a)}: Modified neural activation function with a tunable linearity parameter $s$. For $s\!>\!1$ (red curve), the range of arguments $u$ where the neuron response is approximately linear is extended, compared to the regular $\tanh$ function (black curve) that corresponds to $s\!=\!1$.
{\bf (b)}: Accuracy of a balanced ($b\!=\!0$), weakly coupled ($w\!=\!0.1$) reservoir computer (blue curve), the statistical measures $F$, $C$ and $\alpha$, as well as the root-mean-square of all neural activations (magenta curve), plotted as functions of the linearity parameter $s$ over ten orders of magnitude. The accuracy is well beyond 0.9 for a remarkably large range of $s$. Quantities $F$, $C$, $\alpha$ and the RMS activation indicate a 'calm' reservoir, with small neural signal amplitudes, due to the small recurrent coupling.
{\bf (c)}: Time-dependent activations in the weakly coupled, balanced reservoir during the circle task. Top: Case of standard $\tanh$ activation functions with $s\!=\!1$. Middle: Case of linear neurons. Bottom:  Difference of the latter two, which is only of order $10^{-3}$.  The panels {\bf (d-g)} show the distributions of different PCA components of the reservoir activations, with classes in color coding, both for linear neurons (left of each pair) and $\tanh$ neurons (right of each pair). The PCA components 0-2 do not allow for linear separation of the classes, and the distributions of these components are very similar for linear and $\tanh$ neurons (d,e). Small differences become visible in PCA component 3. For the $\tanh$ neurons, the distributions of the two classes are slightly shifted along this axis (f, right plot of the pair). Drastic differences between the neuron types appear in the combination of PCA components 3 and 4. Here, the classes are linear separable for $\tanh$ neurons, while they completely overlap for linear neurons.  Together, these results indicate that the effect of nonlinearity is crucial for RC performance, but very subtle. 
}
\label{fig_TuneNonLin}
\end{figure}

\NI Returning to $\tanh$-like neurons, we next explore which degree of nonlinearity is required to achieve a high accuracy in the circle problem. For this purpose, we use a tunable activation function $y=s\cdot\tanh(u/s)$, in which the quasi-linear range around $u\!=\!0$ can be extended (or compressed) by a 'linearity parameter' $s$ (Fig.\ref{fig_TuneNonLin}(a)). For $s\!=\!1$, we obtain the ordinary $\tanh$ function (black curve), which 'appears quite linear' in the range $u\in\left[-1,+1\right]$, but for larger arguments $u$ bends away from $y=u$ to approach the asymptotic saturation values $y\rightarrow\pm1$. For $s>1$, the approximately linear range of the activation function correspondingly extends to $u\in\left[-s,+s\right]$. 

\NI When we compare the performance of RCs with tunable neural nonlinearity in the circle task (Fig.\ref{fig_TuneNonLin}(b), blue curve), we find that the accuracy is actually best for  linearity parameters around $s_{opt}\approx 0$. For considerably smaller values of $s$, the accuracy drops, eventually reaching a kind of plateau for $s\!<\!0.01$. The accuracy also drops for $s\!\gg\!s_{opt}$, however rather slowly. Even for $s\approx 100$, where a plot of the activation function appears perfectly linear over the range of typical $u$-values, the accuracy is still around $0.9$. 

\NI One might suspect that even such a small degree of nonlinearity in the activation function around $u\!\approx\!0$ can lead to positive feedback loops, which eventually may drive the reservoir signals into the strongly nonlinear saturation regime. However, this is not the case, as the root-mean-square average of all activations (pooling over times and neurons) remains smaller than $0.2$ over the whole range of considered linearity parameters $s$ (Fig.\ref{fig_TuneNonLin}(b, magenta curve).

\NI Together, these results demonstrate that due to the ability of the readout-layer to sensitively pick out task-relevant information from the reservoir, even an extremely small amount of nonlinearity is sufficient.

\subsection{Signatures of nonlinear data processing}

\NI Considering that a given reservoir with $\tanh$ neurons enables a close to perfect performance in the circle task, while the very same network with linear neurons produces only chance level accuracy (Compare Sec.\ref{cmpLinTanh}), we next investigate how the nonlinearity manifests in the time-dependence and statistics of the neural activations. 

\NI For this purpose, we again analyze the weakly coupled ($w\!=\!0.1$), balanced ($b\!=\!0$) reservoir with $N\!=\!10$ neurons, continuously fed with data from the circle task. In two independent runs, one for linear and one for standard $\tanh$ neurons, we compute the time dependent activations of the reservoir, starting from the same random initial condition and using precisely the same input data set (Fig.\ref{fig_TuneNonLin}(c-g). 

\NI A direct comparison of the activations (panel c) show that the differences between reservoirs with $\tanh$ and linear neurons are extremely small - in our case around 100 times smaller than the signals themselves. It is therefore highly surprising that such small differences can be exploited by the readout layer to classify circle data in the $\tanh$ case.

\NI To convince ourselves that there are indeed subtle but exploitable differences between the $\tanh$ and linear activations, we next apply Principal Component Analysis (PCA). For this purpose, the distribution of all momentary reservoir state vectors $\mathbf{y}$ during the simulation run is considered as a point cloud in 10-dimensional space. PCA then finds the mutually orthogonal axes of maximal variation for this cloud, ordering the indices of the axes with respect to their degree of data variability. 

\NI When analyzing the joint distributions of the 'main' PCA components 0 and 1 (Panels d), which carry most of the reservoir state variation, we still find the circular symmetry of the original input data. Moreover, there are hardly any differences visible between linear and $\tanh$ neurons.

\NI In the joint distribution of PCA components 1 and 2 (Panels e), the symmetry has changed, but it would still not be possible to linearly separate the two classes based on these components. Again, no differences are apparent between the two types of neurons. 

\NI Combining PCA components 2 and 3 (Panels f), a slight shift of the the two classes along axis 3 appears for the $\tanh$ neurons, a feature which may be exploitable for classification.

\NI Finally, combining PCA components 3 and 4 (Panels g), a drastic difference appears between the two neuron types: While the projected classes fully overlap for the linear neurons, they are now linearly separable for the $\tanh$ neurons. 

\NI We conclude that the nonlinearity of the $\tanh$ neurons only leads to very subtle changes in the time evolution of the reservoir states. However, even such subtle changes are reliably exploited by the readout layer, thereby enabling perfect classification in the case circle task.
 
\subsection{Reservoir dynamics and RC performance}

In the past sections on RC performance, we have mainly focused on weakly coupled and balanced reservoirs, because they offer 'calm' working conditions, with no performance-degrading interference between spontaneous reservoir dynamics and input-related data processing. 

\begin{figure}[ht!]
\centering
\includegraphics[width=0.9\linewidth]{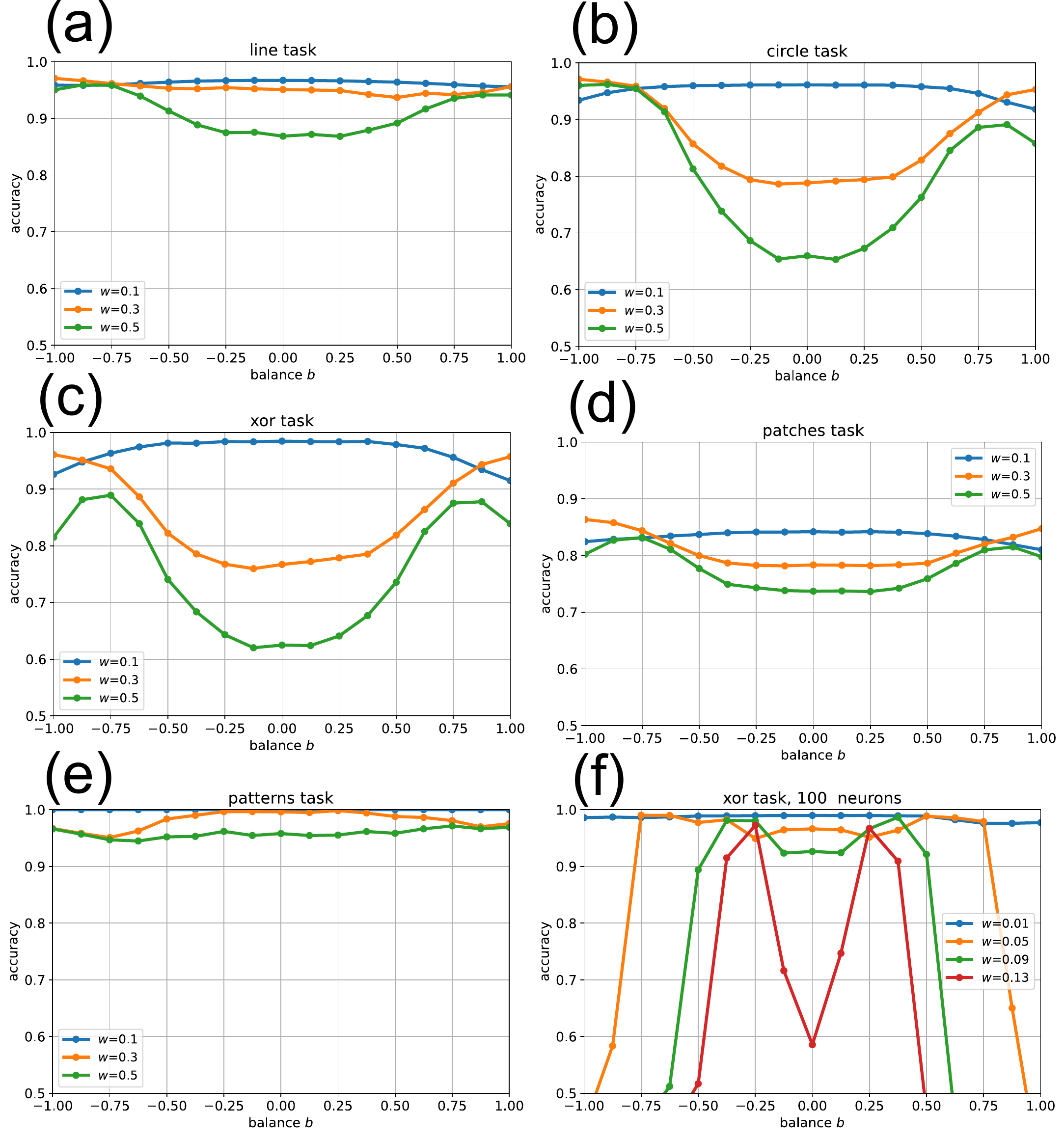}
\caption{{\bf Effect of reservoir dynamics on RC performance.} Panels {\bf (a)-(e)} show the accuracy of a RC (reservoir of $N\!=\!10$ neurons with $\tanh$ activation) in different tasks, as functions of the excitatory/inhibitory balance $b$, for three different recurrent coupling strengths $w\!=\!0.1$ (weak coupling), $w\!=\!0.3$ (medium coupling) and $w\!=\!0.5$ (strong coupling). Plotted accuracies are averaged over 1000 random reservoirs with the same statistical control parameters $w$ and $b$. For balanced systems ($b\!\approx\!0$), consistently for all considered tasks, we find that the accuracy only degrades as the coupling $w$ of the reservoir is increased. This effect, which is particularly pronounced in the 'circle' task (b) and in the 'XOR' task (c), is attributed with the onset of chaotic spontaneous dynamics of the reservoir, which corrupts the task-related information processing. In the extreme oscillatory ($b\!\approx\!-1$) and fixed point ($b\!\approx\!+1$) regimes, the coupling strength $w$ has a much weaker influence on the accuracy, compared to the chaotic regime ($b\!\approx\!0$). In the 'XOR' (c) task (and similarly also in some of the other tasks), the accuracy of the strongly coupled system ($w\!=\!0.5$, green lines) shows two peaks as a function of the balance, located at around $b\!\approx\!\pm0.75$. In these specific cases, the performance of the RC is indeed best at the 'edges of chaos'. Panel {\bf (f)} again shows the accuracy versus balance in the XOR task, however with a reservoir of $N\!=\!100$ neurons and for recurrent couplings $w$ that have been adapted according to the network size. The appearance, at larger couplings, of two peaks at the edges of chaos is even more pronounced in these larger systems. 
}
\label{fig_TuneBalance}
\end{figure}

\NI Next, we investigate how the accuracy in our five types of classification tasks depends on the recurrent coupling strength $w$ and on the excitatory/inhibitory balance $b$, the two key parameter that control the dynamical regime of the reservoir. Using the 10-neuron reservoir with standard $\tanh$ neurons, we again consider the three selected cases of weak coupling $w\!=\!0.1$, medium coupling $w\!=\!0.3$ and strong coupling $w\!=\!0.5$, while the balance is continuously scanned from $b=-1$ to $b=+1$, thus covering the (more or less pronounced) oscillatory, chaotic and fixed point regimes (Fig.\ref{fig_TuneBalance}).

\NI For balanced systems ($b\!\approx\!0$), consistently for all considered tasks, we find that the accuracy only degrades as the coupling $w$ of the reservoir is increased. This effect, which is particularly pronounced in the 'circle' task (b) and in the 'XOR' task (c), can be attributed to the onset of chaotic spontaneous dynamics of the reservoir, which corrupts the task-related information processing. 

\NI In the extreme oscillatory ($b\!\approx\!-1$) and fixed point ($b\!\approx\!+1$) regimes, the coupling strength $w$ has generally a much weaker influence on the accuracy, compared to the chaotic regime ($b\!\approx\!0$). Remarkably, even in the medium and strongly coupled cases, where the spontaneous reservoir dynamics is strongly oscillatory at $b\!\approx\!-1$, the accuracy remains well above $0.9$ for most tasks (apart from 'XOR' and 'patches'). Here, the task-related computations are just tiny perturbations on top of the oscillations, but the readout-layer nevertheless can make use of them.

\NI Strikingly, in the 'XOR' (c) task - and similarly also in some of the other tasks - the accuracy of the strongly coupled system ($w\!=\!0.5$, green lines) shows two peaks as a function of the balance, located at around $b\!\approx\!\pm0.75$. At least in these specific cases, the performance of the RC is indeed best at the 'edges of chaos'. The two peaks get even more pronounced when the number of neurons in the reservoir is increased from $N\!=\!10$ to $N\!=\!100$ (f).

\section{Discussion}

\NI In this work, we have used a reservoir computer with $\tanh$-neurons to investigate  the interplay of spontaneous and input-induced reservoir dynamics, the effects of nonlinearity, and the resulting performance in simple classification tasks.

\paragraph{Tuning reservoir dynamics by weight statistics:} Based on a combination of measures that together characterize the state of an RNN, we have shown how the dynamical behavior of a reservoir can be controlled by two key parameters, the coupling strength $w$ and the excitatory/inhibitory balance $b$. In the weak coupling regime at small $w$, the reservoir is quiet, with neural activations resting at fixed values determined by the random biases. By contrast, in the medium and strong coupling regimes, the reservoir shows spontaneous dynamics, which can be either oscillatory, chaotic, or fixed-point-like, depending on the balance $b$. Feeding input signals of reasonable strength into the reservoir does not have a strong effect on the reservoir dynamics, particularly in the strong coupling regime: When the reservoir is - for example - tuned into the fully oscillatory phase, the input-related perturbations merely appear as 'ripples superposed on huge neural activation waves', but nevertheless the readout layer can exploit these delicate perturbations to classify data with an accuracy significantly above chance level.  

\paragraph{Nonlinearity is necessary but only to a minimal degree:} We have confirmed that even an isolated readout layer can classify linearly separable data, but that it needs the help of a nonlinear reservoir for handling more complex data distributions. In artificial neural networks, nonlinearity is often associated with neurons being driven deeply into their saturation regimes, where outputs become constrained to the extremes of the activation functions. However, our results demonstrate that even when the reservoir is in a quiet global state, with all neurons operating close to zero activation and far from saturation, higher-order terms in the Taylor expansion of the activation function already provide sufficient nonlinearity to perform complex tasks. Indeed, the effects of this weak nonlinearity are so subtle that they become apparent only in the higher order principal components of the time-dependent reservoir state - components that contribute only very little to the overall variance of the high-dimensional state distribution and thus might be misinterpreted as irrelevant noise. These findings may point to a new, weakly nonlinear operating regime for reservoir computers, although future work still needs to determine the full range of tasks that can be solved in that way.

\paragraph{Performance peaks at the edges of chaos, but weak coupling works even better:} Consistent with the long-standing "edge of chaos" hypothesis, we observed that reservoirs with medium or strong coupling perform optimally near the boundary between chaotic and ordered dynamics. However, at the same time our findings revealed that reservoirs with weak coupling, where all spontaneous dynamics is suppressed from the outset, and where also the input signals do not significantly disturb this quiet global state, can perform even better in the tasks investigated here. This weak coupling regime is attractive from an engineering point of view, because it combines high computational power with simpler internal dynamics. 

\paragraph{Weakly coupled, weakly nonlinear reservoir computing:} Our experiments raise an important question for the field: why aim for strong nonlinearity and complex dynamics when minimal configurations appear sufficient? Our results suggest that even with low degrees of nonlinearity and coupling, reservoirs can generate rich enough internal representations to solve a wide range of tasks. This shift toward simplicity could redefine the approach to designing reservoir computing systems, emphasizing effectiveness over dynamical complexity. Such a paradigm could be particularly beneficial for scaling reservoir computing systems to larger networks or for applications that require stability and robustness without sacrificing performance.

\paragraph{Broader implications and biological relevance:} The discovery of the weakly coupled and weakly nonlinear operating regime for artificial reservoirs has possible implications for biological neural systems as well. The brain, as a highly efficient and adaptable computational network, may rely on similar principles, balancing stability with sufficient flexibility for information processing. On the one hand, biological neural networks might exploit minimal nonlinearity in a manner similar to our simple model systems, achieving complex computations without driving the network into a regime where all neurons fire at maximum rate most of the times. On the other hand, our finding that low-amplitude, task-related fluctuations of the network's state can 'surf' on high-amplitude, spontaneous activation waves, points to other intriguing possibilities: Regular waves of reservoir activity - a very prominent feature in the brain - may alternately drive the system through weakly and strongly nonlinear regimes of the neural response functions, leading correspondingly to different representations of the same task-related information. Subsequent processing stages, analogous to a readout layer, may then select only a specific temporal section from this ongoing stream of alternate representations - the one which is most useful for the task at hand. This phase-locked, time-distributed way of information processing would provide an intriguing new interpretation of the well-studied 'brain waves'. As a consequence, interesting information may actually be hidden within the higher-order principal components of brain signals, obtained from EEG and local field potential measurements.

\section{Additional Information}

\subsection{Author contributions}

CM conceived the study, implemented the methods, evaluated the data, and wrote the paper. PK conceived the study, discussed the results, acquired funding and wrote the paper. AS discussed the results and acquired funding. AM discussed the results and provided resources.

\subsection{Funding}
This work was funded by the Deutsche Forschungsgemeinschaft (DFG, German Research Foundation): KR\,5148/3-1 (project number 510395418), KR\,5148/5-1 (project number 542747151), and GRK\,2839 (project number 468527017) to PK, and grant SCHI\,1482/3-1 (project number 451810794) to AS.

\subsection{Competing interests statement}
The authors declare no competing interests.

\subsection{Data availability statement}
The complete data and analysis programs will be made available upon reasonable request.

\subsection{Third party rights}
All material used in the paper are the intellectual property of the authors.


\newpage

\newpage

\newpage


\newpage


\newpage


\newpage


\end{document}